\documentclass[]{spie}  

\usepackage{amsfonts}
\usepackage{amsmath}

\usepackage{amsmath,amsfonts,amssymb}
\usepackage{graphicx}
\usepackage[colorlinks=true, allcolors=blue]{hyperref}

\title{Variational Inference for Quantifying Inter-observer Variability in Segmentation of Anatomical Structures}

\author[a]{Xiaofeng Liu}
\author[a]{Fangxu Xing}
\author[a]{Thibault Marin}
\author[a]{Georges El Fakhri}
\author[a]{Jonghye Woo}

\affil[a]{Gordon Center for Medical Imaging, Massachusetts General Hospital and Harvard Medical School, Boston, MA 02114 USA}

\begin{document} 
\maketitle

\begin{abstract}
Lesions or organ boundaries visible through medical imaging data are often ambiguous, thus resulting in significant variations in multi-reader delineations, i.e., the source of aleatoric uncertainty. In particular, quantifying the inter-observer variability of manual annotations with Magnetic Resonance (MR) Imaging data plays a crucial role in establishing a reference standard for various diagnosis and treatment tasks. Most segmentation methods, however, simply model a mapping from an image to its single segmentation map and do not take the disagreement of annotators into consideration. In order to account for inter-observer variability, without sacrificing accuracy, we propose a novel variational inference framework to model the distribution of plausible segmentation maps, given a specific MR image, which explicitly represents the multi-reader variability. Specifically, we resort to a latent vector to encode the multi-reader variability and counteract the inherent information loss in the imaging data. Then, we apply a variational autoencoder network and optimize its evidence lower bound (ELBO) to efficiently approximate the distribution of the segmentation map, given an MR image. Experimental results, carried out with the QUBIQ brain growth MRI segmentation datasets with seven annotators, demonstrate the effectiveness of our approach.
\end{abstract}


\section{Introduction}

Segmentation has been a crucial prerequisite to provide a basis for various downstream diagnosis and treatment tasks \cite{liu2022self}. Lesions or organ boundaries visible through medical imaging data, however, are not perfectly identified in many cases, because of imaging related artifacts or characteristics of lesions, and therefore multimodal imaging approaches are often used to accurately characterize lesions or organ boundaries~\cite{marin2021deep,joskowicz2019inter}. For example, different tumor tissues can be observed with different MRI modalities; it is challenging to clearly delineate the anatomical and pathological structures with a single MR image \cite{becker2019variability}. In practice, the ambiguity can result in large variations in manual delineations from different observers, thereby leading to aleatoric uncertainty in training \cite{liu2021generative}.

Typical segmentation solutions simply train a network to map an image to its segmentation label, which does not consider the variability of multiple labels. Using only a single most likely segmentation map, however, could lead to misdiagnosis and sub-optimal treatment \cite{joskowicz2019inter}. Therefore, quantifying the inter-observer variability of multiple annotators has been an important task in establishing a reference standard for clinical decision-making \cite{becker2019variability}. With the distribution of possible segmentation maps, the underlying aleatoric uncertainty can be quantified in each voxel, and further diagnostic tests can be planned to resolve the ambiguities inherent in the segmentation result \cite{becker2019variability}. In addition, clinicians can select the appropriate segmentation maps from this distribution for subsequent tasks.

Recently, the QUBIQ challenge database\footnote{https://qubiq.grand-challenge.org/} was proposed to benchmark the segmentation algorithms, returning inter-observer variability estimations. The top two methods \cite{maestimating,yangintegrated} simply used the labels from different annotators to train the neural networks independently, and averaged the predictions of these networks as the inter-observer variability measurement. With insufficient labeled training data, however, the epistemic uncertainty w.r.t. the model parameters can be significant \cite{der2009aleatory,kendall2017uncertainties,liu2021generative}, and the resulting uncertainty map can be a mixture of aleatoric and epistemic uncertainties \cite{der2009aleatory,kendall2017uncertainties}. In addition, the training with multiple annotations can potentially be used to improve segmentation accuracy. Furthermore, Bayesian dropout  \cite{liu2021generative} can be an alternative to measure the aleatoric uncertainty, while its simple Bernoulli prior in the dropout operation may limit its representability.

In this work, we propose a novel variational inference framework to explicitly model the inter-observer variability, i.e., the posterior distribution of segmentation maps given an image. All of the labels are used to train a network jointly, which can potentially improve the accuracy and eliminate the epistemic uncertainty. 


\begin{figure}[t]
\begin{center}
\includegraphics[width=1\linewidth]{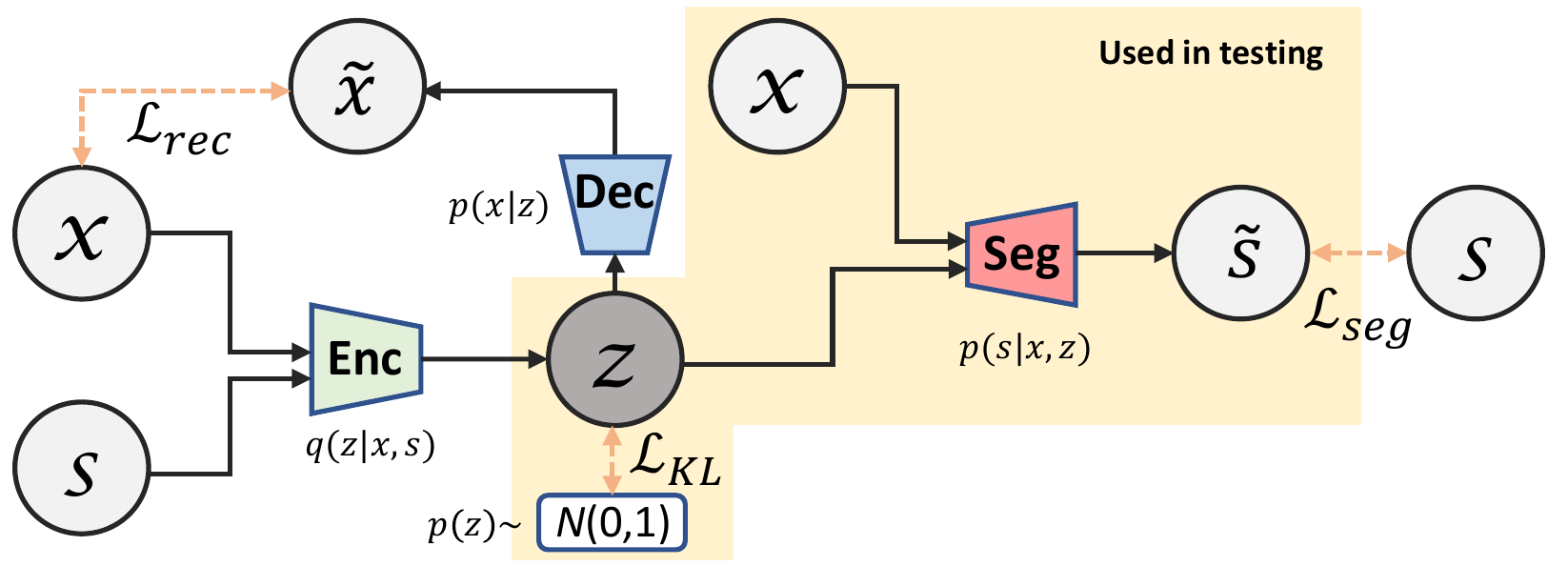}
\end{center} 
\caption{Illustration of our variational inference framework for quantifying inter-observer variability, which consists of an encoder (Enc), a decoder for reconstruction (Dec), and a segmentation network (Seg). Observed variables are shaded in gray. Only the yellow shaded part is used in testing. } 
\label{ccc}\end{figure}

\section{Related work}

Segmentation has been a vital task in medical image analysis \cite{liu2020importance,liu2020severity,liu2020wasserstein,liu2021segmentation,wang2021automated,liu2020reinforced}. The semantic segmentation makes pixel-wise classification and provides more explainable results for the subsequent decision making \cite{liu2021energy,liu2021Off-the-Shelf}. However, the conventional training of segmentation neural networks utilized a single label as the ground-truth, and did not consider the inter-observer variability. 

The classical solution to consider the variability of multiple label is to use the probabilistic graphical models, e.g., conditional random fields have been proposed to measure the joint probability distribution over labels \cite{blake2011markov}, which relies on the maximum a posteriori estimate \cite{monteiro2020stochastic}. Following the development of deep learning, conditional variational autoencoders have been utilized to 
model the spatially correlated aleatoric uncertainty \cite{kohl2018probabilistic,kohl2019hierarchical,baumgartner2019phiseg}. Hu et al. \cite{hu2019supervised} extended this framework, by regressing the uncertainty maps in a supervised manner. In this work, we propose a new perspective of the variational inference along with a novel formulation of its ELBO.

\section{METHODS}

In the task of multi-reader delineations, each image $x$ has several segmentation labels $s$ from different annotators. To quantify the inter-observer variability, we need to model the conditional distribution of $p(s|x)$. Directly modeling $p(s|x)$ with a forward model is intractable, since the segmentation of $x\rightarrow s$ can be inherently ambiguous and ill-posed, due to the insufficient information in imaging data.~In this work, we aim to approximate $p(s|x)$ w.r.t. the Kullback-Leibler (KL) divergence, which is equivalent to maximizing its log-likelihood $\log p(s|x)$ \cite{odaibo2019tutorial}. We resort to a latent vector $z$ to encode the multi-reader variability and counteract the inherent information loss. With the latent variable $z$, we are able to use an ELBO for $\log p(s|x)$, which can be efficiently estimated and optimized as follows: 
\begin{align}\label{222}
\begin{matrix}\log p(s|x)=\underbrace{KL[q(z|x,s)||p(z|x,s)] - \log p(x)}  \\ {\geq 0} \end{matrix}~~~~~~~~~~~~~~~~~~~~~~~~~~~~~~~~~~~~~~~~~~ \nonumber\\ 
\begin{matrix}\underbrace{+^{\mathbb{~~~~~E}}_{z\sim q(z|x,s)}\log p(s|x,z) + ^{\mathbb{~~~~~E}}_{z\sim q(z|x,s)}\log p(x|z) - KL[q(z|x,s)||p(z)],}\\ {ELBO} \end{matrix}  
\end{align} where $KL[q(z|x,s)||p(z|x,s)]$ and $- ^{\mathbb{~~~~~E}}_{z\sim q(z|x,s)}\log p(x)$ are the non-negative terms, and the last three terms can be regarded as the evidence lower bound. We note that KL divergence is non-negative, and $p(x)\in[0,1]$. Maximizing the ELBO can efficiently maximize $\log p(s|x)$, i.e., learning to approach the underlying distribution of $p(s|x)$ using the segmentation network.

\noindent\textbf{Proof.} We have the following decomposition to derive the ELBO of $\log p(s|x)$:  
\begin{align}\label{111} 
KL[q(z|x,s)||p(z|x,s)] &= \sum_{z|x,s} q(z|x,s)\log \frac{q(z|x,s)}{p(z|x,s)} \nonumber\\&= \sum_{z|x,s} q(z|x,s)\{\log {q(z|x,s)} - \log {p(z|x,s)}\} \nonumber\\
                      &= \sum_{z|x,s} q(z|x,s)\{\log {q(z|x,s)} - \log \frac{p(s|x,z)p(z|x)}{p(s|x)}\} \nonumber\\
                      &= \sum_{z|x,s} q(z|x,s)\{\log {q(z|x,s)} - \log p(s|x,z) + \log p(s|x) - \log p(z|x)\} \nonumber\\
                      &= \sum_{z|x,s} q(z|x,s)\{\log {q(z|x,s)} - \log p(s|x,z) + \log p(s|x) - \log p(z) - \log p(x|z) + \log p(x)\}  ~\footnotemark \nonumber\\  
                      &= KL[q(z|x,s)||p(z)]- ~^{\mathbb{~~~~~E}}_{z\sim q(z|x,s)}\log p(s|x,z) + \log p(s|x) - ~^{\mathbb{~~~~~E}}_{z\sim q(z|x,s)}\log p(x|z) + \log p(x)
\end{align}\footnotetext{Can be rewritten as $\sum_{z|x,s} q(z|x,s)\log\frac{q(z|x,s)}{p(z)} - \sum_{z|x,s} q(z|x,s)\log p(s|x,z) + \sum_{z|x,s} q(z|x,s)\log p(s|x) - \sum_{z|x,s} q(z|x,s)\log p(x|z) + \sum_{z|x,s} q(z|x,s)\log p(x)$.\vspace{+5pt}}where $KL[\cdot]$ indicates the KL divergence. We note that, with the Baye's rule, we have $p(z|x,s)=\frac{p(s|x,z)p(z|x)}{p(s|x)}$\footnote{$p(z|x,s)=\frac{p(z,x,s)}{p(x,s)}=\frac{p(s|x,z)p(x,z)}{p(s|x)p(x)}=\frac{p(s|x,z)p(z|x)p(x)}{p(s|x)p(x)}=\frac{p(s|x,z)p(z|x)}{p(s|x)}$}, and $p(z|x)=\frac{p(x|z)p(z)}{p(x)}$. In addition, $\log p(s|x)$ 
and $\log p(x)$ are independent of the value of $z\sim q(z|x,s)$.

Based on the ELBO, we construct our variational inference framework as shown in Fig. \ref{ccc}. The encoder takes both $x$ and $s$ as input to model $q(z|x,s)$ and outputs the latent vector $z$. In addition, we require $z$, following a simple prior distribution, which is usually a Gaussian distribution. The decoder takes $z$ as input to reconstruct $x$, which models the distribution of $p(x|z)$. The segmentation network takes both $x$ and $z$ as input, and the segmentation can be different w.r.t. different values of $z$.

Empirically, we maximize $^{\mathbb{~~~~~E}}_{z\sim q(z|x,s)}\log p(s|x,z)$ by minimizing the cross-entropy (CE) loss of segmentation $\mathcal{L}_{seg}$. All of the samples are trained with multiple labels. Maximizing $^{\mathbb{~~~~~E}}_{z\sim q(z|x,s)}\log p(x|z)$ can be achieved by minimizing the reconstruction loss $\mathcal{L}_{rec}$ as in vanilla VAEs. We simply adopt the pixel-wise mean square error (MSE). Suppose that $\tilde{{x}}$ is the reconstructed ${x}$, then their reconstruction loss can be formulated as: 
\begin{align}\label{eq:m2}
    \mathcal{L}_{rec}({x},\tilde{{x}})=\frac{1}{2}||{x} -\tilde{{x}} ||^2_2.
\end{align}
In addition, maximizing $-KL[q(z|x,s)||p(z)]$ is equivalent to minimizing $KL[q(z|x,s)||p(z)]$, which enforces the encoded latent vectors to align with its prior. We adopt the reparameterization trick to formulate the KL divergence. Similar to vanilla VAEs \cite{kingma2013auto,liu2020auto3d,liu2021unified}, the encoder has two output vectors, $i.e.,$ $\mu$ and $\sigma$. We then utilize the reparameterization trick ${z}=\mu+\sigma\odot\epsilon$, where $\epsilon\in \mathcal{N}(0,I)$. The posterior distribution of ${z}$ is $q({z}|{x},{y})\sim \mathcal{N}({z};\mu,\sigma^2)$. Specifically, the KL-divergence can be computed as 
\begin{align}  \label{eq:m1}
   \mathcal{L}_{KL}({z};\mu,\sigma)=\frac{1}{2}\sum^{N}_{j=1}(1+{\rm log}(\sigma_j^2)-\mu_j^2-\sigma_j^2),
\end{align}where $N=6$ is the dimension of the latent vector ${z}$.

In testing, only the segmentation network is used. We sample on the prior distribution of $p(z)$, i.e., the Gaussian distribution, to have different $z$, and generate different segmentation predictions. After the training, the network can model $p(s|x,z)$, and $z$ explicitly encodes the inter-observer variability.

\section{RESULTS}

The QUBIQ brain growth MRI segmentation database consists of a total of 34 subjects for training and a total of 5 subjects for validation.~Each MRI slice has seven segmentation annotations. We note that the testing set is private, and the validation is used for comparison. Given a single image $x$, we were able to output the distribution of $s$ with the sampled different $z$, rather than a deterministic output in conventional segmentation networks. We used the same 2D U-Net backbone for the segmentation network as in ESULR \cite{maestimating}. The statistic of the brain growth sub-task on QUBIQ is shown in Table \ref{tabel:0}.

\begin{figure}[t]
\begin{center}
\includegraphics[width=1\linewidth]{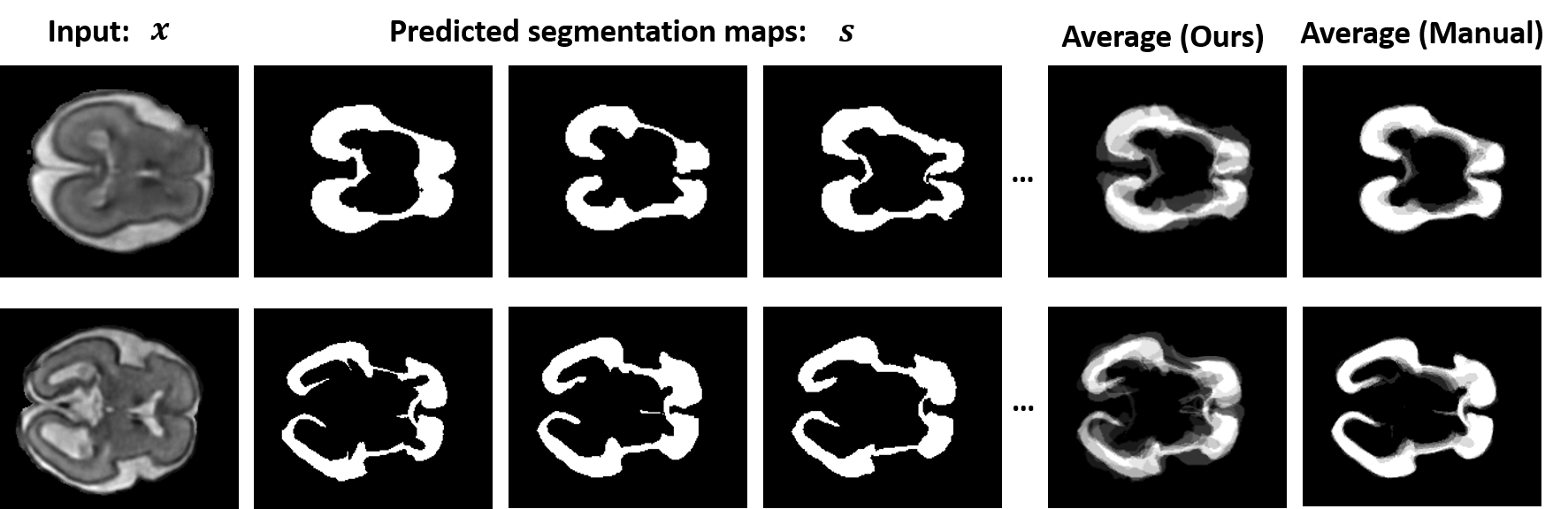} 
\end{center} 
\caption{Illustration of the predicted segmentation distribution and the comparison with the averaged manual labels. The averaged segmentation map is regarded as the ground truth quantified by inter-observer variability on QUBIQ. }\vspace{+5pt}
\label{ddd}\end{figure}

Following the standard evaluation metrics on QUBIQ, we averaged 7 predictions as the confidence map and compared them with the average of seven manual labels with the continuous Dice similarity score. We note that the averaged value of each pixel is in $[0,1]$. The confidence map can be regarded as the reciprocal of the aleatoric uncertainty. Fig.~\ref{ddd} depicts representative outputs of a validation sample and the averaged inter-observer variability results. The numerical results are shown in Table~\ref{tabel:1}. Our framework outperformed the independent training methods \cite{maestimating,yangintegrated} and the Bayes dropout used in \cite{liu2021generative} for the aleatoric uncertainty quantification. 

\begin{figure}[t]
\begin{center}
\includegraphics[width=1\linewidth]{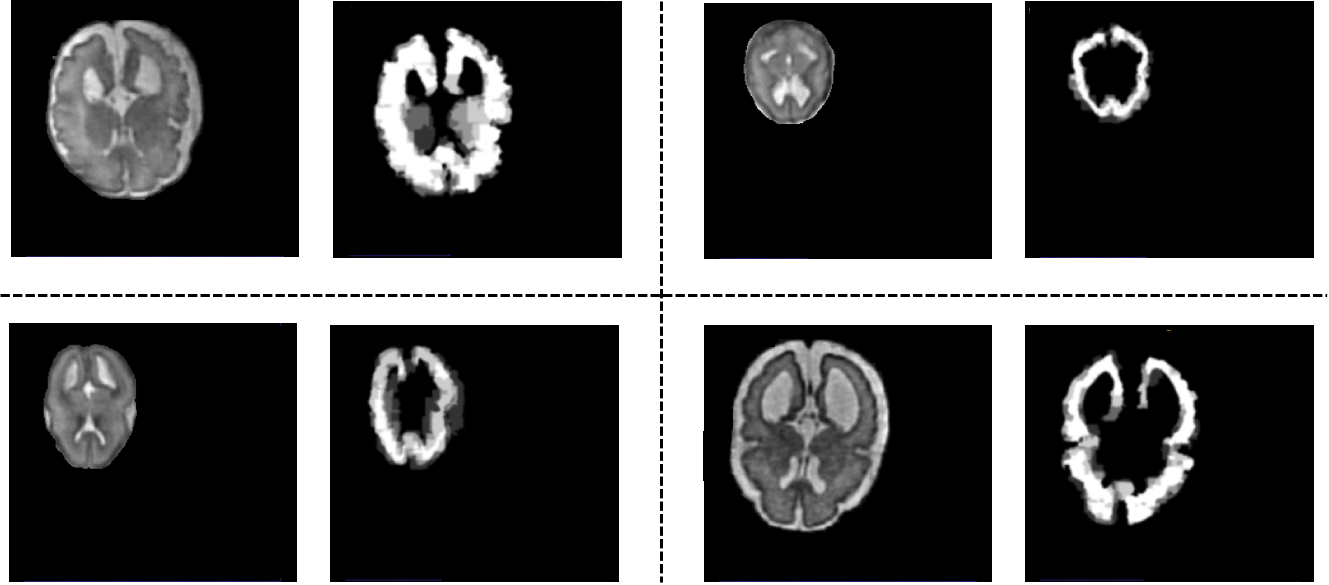} 
\end{center} 
\caption{Illustration of the MR slice and its corresponding predicted segmentation distribution in QUBIQ testing set. }\vspace{+10pt}
\label{eee}\end{figure}

In Figure \ref{eee}, we show the uncertainty estimation in the testing set. We note that there is no publicly available ground truth for the testing data. The to-be segmented parts vary significantly w.r.t.~the shape, scale, and location. It can be quite challenging to precisely delineate the anatomical structures with limited training data. We can see that our predictions align well with the shape. 

\begin{table}[t]
\centering\label{tabel:0} \vspace{+10pt}
\caption{Dataset statistics of the brain growth task on QUBIQ.}\vspace{+5pt}
\resizebox{0.65\linewidth}{!}{
\begin{tabular}{c|c|c|c|c|c}
\hline
Sub-task & Training & Validation & Testing & Task & Channels\\\hline
brain-growth & 34 &5 &10 &1 &1\\\hline
\end{tabular}
}
\end{table}

\section{CONCLUSION}

In this work, we proposed a novel variational inference framework to estimate the posterior distribution of multi-reader segmentation labels from a Bayesian's perspective. The derivation of the ELBO of the conditional log-likelihood is analyzed, which can be efficiently estimated and optimized. Then, the networks and loss functions are designed to maximize the ELBO. After training, the segmentation network yielded diverse segmentations with the sampled latent vectors, which inherit the multi-reader variability. All of the labels can be used to train the segmentation networks, instead of independent training. The quantified inter-observer variability, i.e., the averaged segmentation, aligns well with the manual average, when evaluated on the QUBIQ datasbase, which indicates that the aleatoric uncertainty can be accurately measured. In this work, we only explored the brain growth task using the QUBIQ challenge database, while the method can be easily generalized for any inter-observer variability quantification tasks, which is subject to our future work.

\begin{table}[t]
\centering\label{tabel:1} 
\caption{Numerical comparisons of different methods. $\pm$ standard deviation is reported over three times training with random initialization.}\vspace{+10pt}
\resizebox{0.65\linewidth}{!}{
\begin{tabular}{c|c}
\hline

Methods & Continuous Dice Score ($\uparrow$: the larger the better)\\\hline
ESULR \cite{maestimating} (Independent Ave)& 0.5042\\ 
ISMA \cite{yangintegrated} (Independent Ave)&  0.4828 \\ 
Bayes dropout \cite{liu2021generative} &  0.5083 $\pm$ 0.0013\\ \hline
Ours & \textbf{0.5372 $\pm$ 0.0011}\\ \hline

\end{tabular}
}
\end{table}

\acknowledgments 

This work is partially supported by NIH R01DC018511, R01CA165221, and P41EB022544.

\bibliography{report} 

\begin{thebibliography}{10}

\bibitem{liu2022self}
Liu, X., Xing, F., Fakhri, G.~E., and Woo, J., ``Self-semantic contour
  adaptation for cross modality brain tumor segmentation,'' {\em ISBI}  (2022).

\bibitem{marin2021deep}
Marin, T., Zhuo, Y., Lahoud, R.~M., Tian, F., Ma, X., Xing, F., Moteabbed, M.,
  Liu, X., Grogg, K., Shusharina, N., et~al., ``Deep learning-based gtv
  contouring modeling inter-and intra-observer variability in sarcomas,'' {\em
  Radiotherapy and Oncology}  (2021).

\bibitem{joskowicz2019inter}
Joskowicz, L., Cohen, D., Caplan, N., and Sosna, J., ``Inter-observer
  variability of manual contour delineation of structures in ct,'' {\em
  European radiology}~{\bf 29}(3),  1391--1399 (2019).

\bibitem{becker2019variability}
Becker, A.~S., Chaitanya, K., Schawkat, K., Muehlematter, U.~J., H{\"o}tker,
  A.~M., Konukoglu, E., and Donati, O.~F., ``Variability of manual segmentation
  of the prostate in axial t2-weighted mri: A multi-reader study,'' {\em
  European journal of radiology}~{\bf 121},  108716 (2019).

\bibitem{liu2021generative}
Liu, X., Xing, F., Stone, M., Zhuo, J., Reese, T., Prince, J.~L., El~Fakhri,
  G., and Woo, J., ``Generative self-training for cross-domain unsupervised
  tagged-to-cine mri synthesis,'' in [{\em International Conference on Medical
  Image Computing and Computer-Assisted
  Intervention}{\nolinebreak\hspace{0.1em}]},   138--148, Springer (2021).

\bibitem{maestimating}
Ma, J., ``Estimating segmentation uncertainties like radiologists,'' {\em
  MICCAI: QUBIQ}  (2020).

\bibitem{yangintegrated}
Yang, Y. and Ma, T., ``Integrated segmentation with multiple annotations,''
  {\em MICCAI: QUBIQ}  (2020).

\bibitem{der2009aleatory}
Der~Kiureghian, A. and Ditlevsen, O., ``Aleatory or epistemic? does it
  matter?,'' {\em Structural safety}~{\bf 31}(2),  105--112 (2009).

\bibitem{kendall2017uncertainties}
Kendall, A. and Gal, Y., ``What uncertainties do we need in bayesian deep
  learning for computer vision?,'' {\em arXiv preprint arXiv:1703.04977}
  (2017).

\bibitem{liu2020importance}
Liu, X., Han, Y., Bai, S., Ge, Y., Wang, T., Han, X., Li, S., You, J., and Lu,
  J., ``Importance-aware semantic segmentation in self-driving with discrete
  wasserstein training.,'' in [{\em AAAI}{\nolinebreak\hspace{0.1em}]},
  11629--11636 (2020).

\bibitem{liu2020severity}
Liu, X., Ji, W., You, J., Fakhri, G.~E., and Woo, J., ``Severity-aware semantic
  segmentation with reinforced wasserstein training,'' in [{\em Proceedings of
  the IEEE/CVF Conference on Computer Vision and Pattern
  Recognition}{\nolinebreak\hspace{0.1em}]},   12566--12575 (2020).

\bibitem{liu2020wasserstein}
Liu, X., Lu, Y., Liu, X., Bai, S., Li, S., and You, J., ``Wasserstein loss with
  alternative reinforcement learning for severity-aware semantic
  segmentation,'' {\em IEEE Transactions on Intelligent Transportation Systems}
   (2020).

\bibitem{liu2021segmentation}
Liu, X., Xing, F., Gaggin, H.~K., Wang, W., Kuo, C.-C.~J., Fakhri, G.~E., and
  Woo, J., ``Segmentation of cardiac structures via successive subspace
  learning with saab transform from cine mri,'' {\em arXiv preprint
  arXiv:2107.10718}  (2021).

\bibitem{wang2021automated}
Wang, J., Liu, X., Wang, F., Zheng, L., Gao, F., Zhang, H., Zhang, X., Xie, W.,
  and Wang, B., ``Automated interpretation of congenital heart disease from
  multi-view echocardiograms,'' {\em Medical Image Analysis}~{\bf 69},  101942
  (2021).

\bibitem{liu2020reinforced}
Liu, X., Zhang, Y., Liu, X., Bai, S., Li, S., and You, J., ``Reinforced
  wasserstein training for severity-aware semantic segmentation in autonomous
  driving,'' {\em arXiv preprint arXiv:2008.04751}  (2020).

\bibitem{liu2021energy}
Liu, X., Hu, B., Liu, X., Lu, J., You, J., and Kong, L., ``Energy-constrained
  self-training for unsupervised domain adaptation,'' in [{\em 2020 25th
  International Conference on Pattern Recognition
  (ICPR)}{\nolinebreak\hspace{0.1em}]},   7515--7520, IEEE (2021).

\bibitem{liu2021Off-the-Shelf}
Liu, X., Xing, F., El~Fakhri, G., and Woo, J., ``Adapting off-the-shelf source
  segmenter for target medical image segmentation,'' in [{\em
  MICCAI}{\nolinebreak\hspace{0.1em}]},  (2021).

\bibitem{blake2011markov}
Blake, A., Kohli, P., and Rother, C.,  [{\em Markov random fields for vision
  and image processing}{\nolinebreak\hspace{0.1em}]}, MIT press (2011).

\bibitem{monteiro2020stochastic}
Monteiro, M., Folgoc, L.~L., de~Castro, D.~C., Pawlowski, N., Marques, B.,
  Kamnitsas, K., van~der Wilk, M., and Glocker, B., ``Stochastic segmentation
  networks: Modelling spatially correlated aleatoric uncertainty,'' {\em arXiv
  preprint arXiv:2006.06015}  (2020).

\bibitem{kohl2018probabilistic}
Kohl, S.~A., Romera-Paredes, B., Meyer, C., De~Fauw, J., Ledsam, J.~R.,
  Maier-Hein, K.~H., Eslami, S., Rezende, D.~J., and Ronneberger, O., ``A
  probabilistic u-net for segmentation of ambiguous images,'' {\em arXiv
  preprint arXiv:1806.05034}  (2018).

\bibitem{kohl2019hierarchical}
Kohl, S.~A., Romera-Paredes, B., Maier-Hein, K.~H., Rezende, D.~J., Eslami, S.,
  Kohli, P., Zisserman, A., and Ronneberger, O., ``A hierarchical probabilistic
  u-net for modeling multi-scale ambiguities,'' {\em arXiv preprint
  arXiv:1905.13077}  (2019).

\bibitem{baumgartner2019phiseg}
Baumgartner, C.~F., Tezcan, K.~C., Chaitanya, K., H{\"o}tker, A.~M.,
  Muehlematter, U.~J., Schawkat, K., Becker, A.~S., Donati, O., and Konukoglu,
  E., ``Phiseg: Capturing uncertainty in medical image segmentation,'' in [{\em
  International Conference on Medical Image Computing and Computer-Assisted
  Intervention}{\nolinebreak\hspace{0.1em}]},   119--127, Springer (2019).

\bibitem{hu2019supervised}
Hu, S., Worrall, D., Knegt, S., Veeling, B., Huisman, H., and Welling, M.,
  ``Supervised uncertainty quantification for segmentation with multiple
  annotations,'' in [{\em International Conference on Medical Image Computing
  and Computer-Assisted Intervention}{\nolinebreak\hspace{0.1em}]},   137--145,
  Springer (2019).

\bibitem{odaibo2019tutorial}
Odaibo, S., ``Tutorial: Deriving the standard variational autoencoder (vae)
  loss function,'' {\em arXiv preprint arXiv:1907.08956}  (2019).

\bibitem{kingma2013auto}
Kingma, D.~P. and Welling, M., ``Auto-encoding variational bayes,'' {\em arXiv
  preprint arXiv:1312.6114}  (2013).

\bibitem{liu2020auto3d}
Liu, X., Che, T., Lu, Y., Yang, C., Li, S., and You, J., ``Auto3d: Novel view
  synthesis through unsupervisely learned variational viewpoint and global 3d
  representation,'' {\em European Conference on Computer Vision}  (2020).

\bibitem{liu2021unified}
Liu, X., Xing, F., El~Fakhri, G., and Woo, J., ``A unified conditional
  disentanglement framework for multimodal brain mr image translation,'' in
  [{\em 2021 IEEE 18th International Symposium on Biomedical Imaging
  (ISBI)}{\nolinebreak\hspace{0.1em}]},   10--14, IEEE (2021).

\end{thebibliography}
\bibliographystyle{spiebib} 

\end{document}